\documentclass[a4paper, 10pt, conference]{ieeeconf}      
\usepackage{FG2026}

\FGfinalcopy 

\IEEEoverridecommandlockouts                              
\usepackage{epsfig} 
\usepackage{times} 
\usepackage{amsmath} 
\usepackage{amssymb}  
\usepackage{graphicx}
\usepackage{booktabs}
\usepackage{multirow} 
\usepackage{makecell}
\usepackage{cite}

\def\FGPaperID{369} 

\title{\LARGE \bf
GaitProtector: Impersonation-Driven Gait De-Identification via \\
Training-Free Diffusion Latent Optimization
}

\author{\parbox{16cm}{\centering
    {\large
    Huiran Duan$^{1*}$,
    Qian Zhou$^{2*}$,
    Zhongliang Guo$^{3}$,
    Junhao Dong$^{4}$,\\
    Yuqi Li$^{1}$,
    Guoying Zhao$^{5,6}$,
    Yingli Tian$^{1\dagger}$}\\
    {\normalsize
    $^{1}$City University of New York, USA\\
    $^{2}$Wuhan University, China\\
    $^{3}$University of Aberdeen, UK\\
    $^{4}$Nanyang Technological University, Singapore\\
    $^{5}$ELLIS Institute Finland\\
    $^{6}$University of Oulu, Finland
    }}
    \thanks{* Equal contribution. $\dagger$ Corresponding author. This work was supported in part by the Research Council of Finland (former Academy of Finland) Academy Professor project EmotionAI (grants 336116, 359894), the University of Oulu \& Research Council of Finland Profi 7 (grant 352788), and EU HORIZON-MSCA-SE-2022 project ACMod (grant 101130271).}
}
\usepackage{fancyhdr}
\begin{document}

\ifFGfinal
\thispagestyle{empty}
\pagestyle{empty}
\else
\author{Anonymous FG2026 submission\\ Paper ID \FGPaperID \\}
\pagestyle{plain}
\fi
\maketitle
 \thispagestyle{fancy}

\begin{abstract}
Conventional gait de-identification methods often encounter an inherent trade-off: they either provide insufficient identity suppression or introduce spatiotemporal distortions that impede structure-sensitive downstream applications. We propose GaitProtector, an impersonation-driven gait de-identification framework that formulates privacy protection as a unified objective with two tightly coupled components: (i) obfuscation, which repels the protected gait from the source identity, and (ii) impersonation, which attracts it toward a selected target identity. The target identity serves as a semantic anchor that biases optimization toward structurally plausible gait patterns under the pretrained diffusion prior, helping preserve dominant body shape and motion dynamics. We instantiate this idea through a training-free diffusion latent optimization pipeline. Instead of retraining a generator for each dataset, we invert each input silhouette sequence into the latent trajectory of a pretrained 3D video diffusion model and iteratively optimize latent codes with a differentiable adversarial objective to synthesize protected gaits. Experiments on the CASIA-B dataset show that GaitProtector achieves a 56.7\% impersonation success rate under black-box gait recognition and reduces Rank-1 identification accuracy from 89.6\% to 15.0\%, while maintaining favorable visual and temporal quality. We further evaluate downstream utility on the Scoliosis1K dataset, where diagnostic accuracy decreases only from 91.4\% to 74.2\%. To the best of our knowledge, this work is the first to leverage pretrained 3D diffusion priors in a training-free manner for silhouette-based gait de-identification.

\end{abstract}

\section{INTRODUCTION}
Gait has emerged as a compelling biometric trait because it can be captured at a distance and in a largely non-cooperative manner~\cite{gait_survey1}, making it particularly attractive for real-world sensing and surveillance scenarios~\cite{gait_survey2}. Across different gait modalities, including RGB videos~\cite{biggait,biggergait}, skeletons~\cite{gaitgraph,gpgait},  SMPL meshes~\cite{gait3d}, and silhouettes~\cite{opengait, gaitgl}, silhouette-based representations have become a mainstream choice. By explicitly discarding background clutter and appearance cues while remaining highly discriminative, silhouettes offer an appealing balance between robustness and efficiency~\cite{gait_silh_survey}.

However, widespread adoption as a recognition modality does not imply that silhouette-based gait data are privacy-safe. In many data-sharing scenarios, silhouettes are often treated as a “sanitized” substitute for RGB videos under the assumption that removing appearance information sufficiently protects identity. As a behavioral biometric, gait encodes identity-linked motion signatures that persist even in silhouette sequences, making identity leakage a realistic risk once such data are released~\cite{li2024sparse}.

To mitigate privacy risks in gait data, prior work has explored various gait de-identification and anonymization strategies that directly transform observed gait sequences, including contour modification~\cite{tieu2017approach}, temporal perturbation~\cite{tieu2019spatio, halder2023gait}, appearance-level editing~\cite{hukkelaas2023deepprivacy2, ma2023pedestrian, ma2024passersby}, and related privacy-preserving identity editing in adjacent domains such as faces~\cite{li2025imperceptible}. Because these methods act on spatial geometry and temporal dynamics that also carry gait structure, stronger transformations often disrupt body geometry or motion consistency~\cite{tieu2017approach, tieu2019spatio}, whereas milder ones leave residual identity-linked patterns exploitable by modern high-capacity gait recognition models~\cite{hirose2023experimental, he2023temporal}. This trade-off is especially problematic for structure-sensitive downstream applications such as clinical gait analysis, where intact trunk structure, posture alignment, and temporal coordination are essential~\cite{Scoliosis1K}.

Motivated by this privacy--utility tension, we formulate gait de-identification as a structured transformation problem. Given a gait silhouette sequence, the goal is to generate a protected sequence that conceals identity information while preserving essential gait structure and dynamics. We therefore combine \emph{obfuscation}, which pushes the protected gait away from the source identity, with \emph{impersonation}, which pulls it toward a target identity. The target identity serves as a semantic anchor under a frozen pretrained 3D video diffusion prior~\cite{VideoFusion,he2022lvdm,blattmann2023align}, biasing latent optimization toward structurally coherent gait patterns and avoiding arbitrary spatial or temporal distortions. In this way, protected gait sequences are synthesized without dataset-specific training or fine-tuning while maintaining structure relevant to downstream applications.

Our contributions are summarized as follows:
\begin{itemize}
    \item We cast gait de-identification as an impersonation-driven process, in which identity protection emerges from the joint effect of obfuscation and impersonation, with impersonation acting as a semantic regularizer that biases transformations toward structurally plausible gait patterns.
    
    \item We propose GaitProtector, a training-free gait de-identification framework that leverages 3D diffusion inversion and latent optimization under a strong generative prior, together with a soft-binarized optimization strategy to effectively handle silhouettes.
    
    \item We conduct a systematic evaluation of gait de-identification that jointly examines privacy protection, visual fidelity on CASIA-B~\cite{CASIA-B}, and downstream utility on Scoliosis1K~\cite{Scoliosis1K}. We further demonstrate robustness under a stronger threat model involving adaptive adversaries performing hard re-binarization. 
\end{itemize}

\section{RELATED WORK}

\subsection{Gait De-Identification Methods}
Existing gait de-identification methods aim to suppress identity information by transforming observed gait representations. Early studies operate directly on silhouette sequences using explicit signal-space manipulations, including phase and shape perturbation~\cite{hirose2019anonymization}, contour deformation~\cite{hirose2022anonymization}, and spatiotemporal warping learned by deep generative models~\cite{tieu2017approach, tieu2019spatio}. Learning-based approaches further train CNN- or GAN-based models to translate input gaits into anonymized outputs, often with additional constraints to preserve pose or motion continuity~\cite{halder2023gait}. Despite differences in parameterization and learning strategy, these approaches share a common design choice: directly modifying the observed gait sequence.

Beyond gait-specific anonymization, privacy protection has also been studied at the appearance level and from adversarial perspectives. Full-body and RGB-based anonymization methods emphasize visual realism and scene consistency through appearance or attribute editing~\cite{hukkelaas2023deepprivacy2, hukkelaas2023realistic, ma2023pedestrian, ma2024passersby,li2025imperceptible}. Adversarial and spoofing studies investigate how sparse or targeted perturbations can manipulate recognition outcomes under strong gait models~\cite{jia2019attacking, he2023temporal, hirose2023experimental, li2024sparse}. Motion privacy has also been studied in skeleton-based settings through selective suppression of identity-related motion cues~\cite{moon2023anonymization}.

Despite these advances, most prior work evaluates privacy protection or visual plausibility in isolation. Joint evaluation of privacy protection, generation quality, and downstream utility remains limited in the gait domain, largely due to the scarcity of task-oriented gait datasets beyond identity recognition. As a result, it is unclear whether identity suppression preserves structurally meaningful gait information required by practical downstream applications. This gap motivates comprehensive privacy–utility evaluation and the use of clinical gait datasets, such as Scoliosis1K, to assess utility beyond recognition tasks.

\subsection{Video Diffusion Models}
Diffusion models have become a dominant approach for video generation by providing strong spatiotemporal priors for realistic motion~\cite{ho2022videodiffusion, VideoFusion,he2022lvdm,blattmann2023align,bar2024lumiere,hacohen2024ltx}. Early studies extend diffusion to the video domain by jointly modeling spatial and temporal denoising processes, demonstrating that diffusion dynamics can naturally enforce temporal coherence~\cite{ho2022videodiffusion, VideoFusion}. Subsequent work improves scalability and efficiency through latent video diffusion. By operating in a compact latent space with dedicated video autoencoders, these methods significantly reduce computational cost while preserving motion fidelity~\cite{he2022lvdm, blattmann2023align}. Large-scale video diffusion models further strengthen these priors by improving temporal compression and long-range consistency, reinforcing pretrained video diffusion models as generic representations of spatiotemporal structure rather than task-specific generators~\cite{bar2024lumiere, hacohen2024ltx}. Recent work also studies deployment-oriented acceleration through quantization and distillation of video diffusion transformers~\cite{feng2025q,feng2025s}.

Alongside generation, diffusion inversion and latent-space manipulation have emerged as a general paradigm for controlled transformation without retraining the generator. Deterministic diffusion formulations such as Denoising Diffusion Implicit Models (DDIM) make inversion tractable by mapping real observations into diffusion trajectories that can be faithfully reconstructed~\cite{song2021ddim}. This capability has been widely adopted in image editing and inverse problems, where pretrained diffusion models are kept fixed and task objectives are enforced through latent optimization or guidance~\cite{meng2022sdedit, mokady2023nulltext}. Recent video editing methods extend this paradigm to the temporal domain by explicitly maintaining inter-frame consistency during inversion and optimization, demonstrating that diffusion priors can support structured video-level manipulation in a training-free manner~\cite{chai2023stablevideo, zhang2025visual}. Together, these advances establish pretrained video diffusion models as fixed spatiotemporal priors for downstream latent-space optimization.

\section{METHOD}
\label{sec:method}

\subsection{Problem Definition and Framework Overview}
\label{sec:method_overview}

We study privacy-preserving generation for silhouette-based gait sequences under a \emph{targeted} de-identification setting. Let a gait silhouette sequence be denoted by $\mathbf{x}\in[0,1]^{L\times H\times W}$, where $L$ is the number of frames and each frame is a silhouette of size $H\times W$. Given a \emph{source} sequence $\mathbf{x}^{(src)}$ with identity $y^{src}$ and a \emph{target} sequence $\mathbf{x}^{(tar)}$ with identity $y^{tar}$, the goal is to synthesize a \emph{protected} sequence $\mathbf{x}^{(pro)}$ that suppresses the source identity and promotes target recognition without compromising perceptual plausibility or downstream utility.

To formalize identity relationships, we consider one or more silhouette-based gait recognition models that map a sequence to an identity embedding space. Let $\mathcal{G}(\cdot)$ denote a gait feature extractor and $\mathcal{D}(\cdot,\cdot)$ a dissimilarity measure in this space. Targeted gait de-identification is characterized by two complementary desiderata: \emph{obfuscation}, which repels the protected sequence from the source identity, and \emph{impersonation}, which attracts it toward the target identity. Accordingly, the identity-level objectives are:
\begin{equation}
\mathcal{L}_{\mathrm{imp}}\big(\mathbf{x}^{(pro)},\mathbf{x}^{(tar)}\big)
=
\mathcal{D}\!\left(\mathcal{G}\!\left(\mathbf{x}^{(pro)}\right),\,\mathcal{G}\!\left(\mathbf{x}^{(tar)}\right)\right),
\label{eq:overview_imp}
\end{equation}
and
\begin{equation}
\mathcal{L}_{\mathrm{obf}}\big(\mathbf{x}^{(pro)},\mathbf{x}^{(src)}\big)
=
-\,\mathcal{D}\!\left(\mathcal{G}\!\left(\mathbf{x}^{(pro)}\right),\,\mathcal{G}\!\left(\mathbf{x}^{(src)}\right)\right),
\label{eq:overview_obf}
\end{equation}
so that minimizing $\mathcal{L}_{\mathrm{imp}}$ encourages proximity to the target identity while minimizing $\mathcal{L}_{\mathrm{obf}}$ encourages separation from the source identity. Beyond identity manipulation, we seek protected outputs that retain the source sequence's dominant body shape and spatiotemporal coherence. In GaitProtector, this plausibility is not enforced through an explicit reconstruction or geometry-preserving penalty; instead, it is encouraged implicitly by the pretrained diffusion prior together with the target-anchored optimization.

\vspace{0.25em}
\noindent\textbf{Training-free formulation.}
GaitProtector is \emph{training-free} in the following sense: it uses (i) a pretrained 3D video diffusion model~\cite{VideoFusion} as a fixed generative prior and (ii) surrogate gait recognition models to provide identity-related guidance, and it produces a protected sample by solving a \emph{pairwise} latent optimization problem for each $(\mathbf{x}^{(src)},\mathbf{x}^{(tar)})$ without training or fine-tuning the generator on a specific dataset. This contrasts with many GAN-based anonymization pipelines that require dataset-specific generator training.

\begin{figure*}[thpb]
  \centering
  \includegraphics[width=0.9\linewidth]{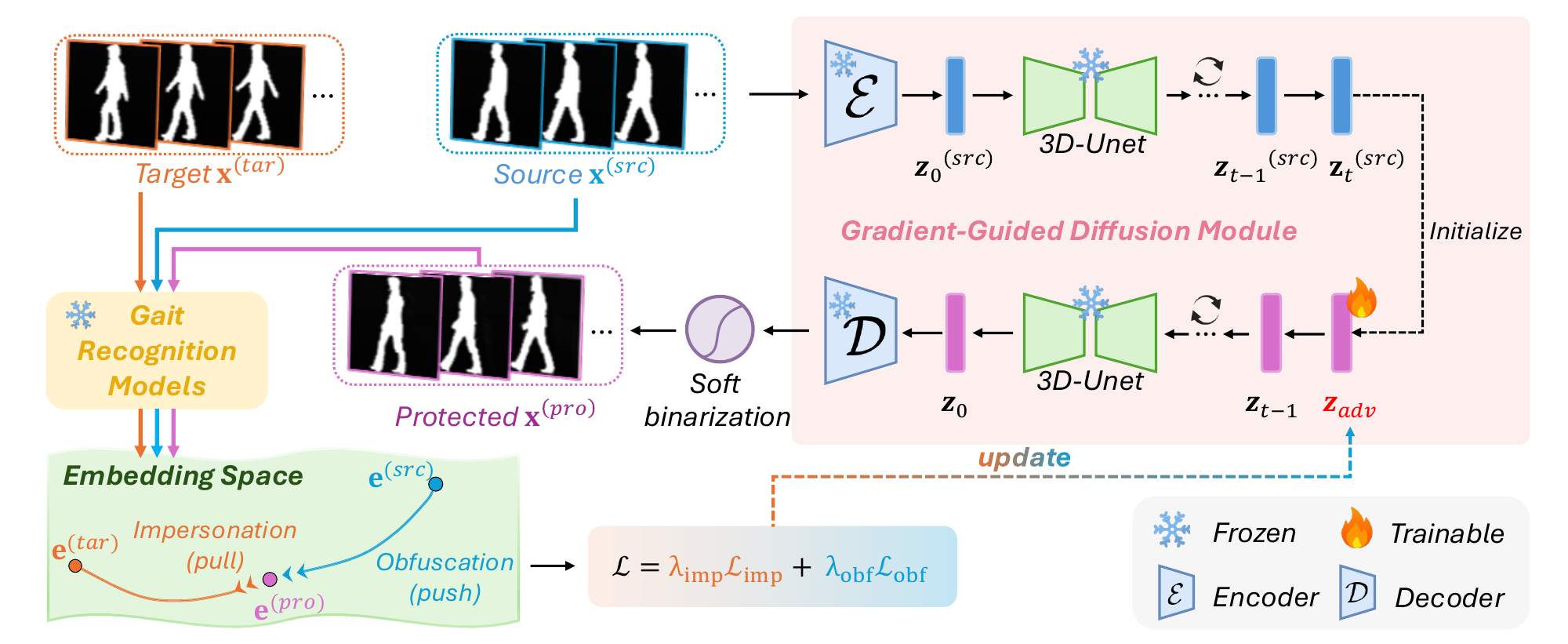}
  \caption{\textbf{Overview of the proposed GaitProtector}. Given a source silhouette sequence $\mathbf{x}^{(src)}$ and a target sequence $\mathbf{x}^{(tar)}$, we first encode $\mathbf{x}^{(src)}$ with a frozen VAE encoder and perform deterministic DDIM inversion to obtain an intermediate latent $\mathbf{z}_t^{(src)}$ at diffusion step $t$. We initialize the optimizable latent variable as $\mathbf{z}_{\mathrm{adv}}\leftarrow \mathbf{z}_t^{(src)}$ and iteratively refine it using gradients from a unified de-identification objective that combines impersonation (attracting $\mathbf{x}^{(pro)}$ toward the target identity) and obfuscation (repelling $\mathbf{x}^{(pro)}$ from the source identity) measured in the embedding spaces of surrogate gait recognition models. At each iteration, $\mathbf{z}_{\mathrm{adv}}$ is deterministically denoised from step $t$ to $0$ (DDIM sampling with $\eta=0$) and then mapped back to the video space by a frozen VAE decoder to obtain a continuous-valued sequence, which is converted to a near-binary silhouette sequence via a differentiable soft binarization. The final protected gait $\mathbf{x}^{(pro)}$ is produced after a fixed iteration budget or convergence.}
  \label{fig:gaitprotector_pipeline}
  \vspace{-0.5em}
\end{figure*}

\vspace{0.25em}
\noindent\textbf{Framework overview.}
Fig.~\ref{fig:gaitprotector_pipeline} summarizes the proposed framework. Starting from $\mathbf{x}^{(src)}$, we encode and invert it into an intermediate diffusion latent at a selected step (Sec.~\ref{sec:method_ddim}), optimize that latent under a unified de-identification objective guided by surrogate recognizers, and generate the final protected sequence by deterministic denoising and decoding. A differentiable soft binarization encourages a bimodal silhouette structure (Sec.~\ref{sec:method_softbin}), and the complete optimization loop is detailed in Sec.~\ref{sec:method_opt}.

\subsection{DDIM Inversion and Intermediate Sampling}
\label{sec:method_ddim}

GaitProtector operates in the latent space of a pretrained 3D video diffusion model. Given a source silhouette sequence $\mathbf{x}^{(src)}$, we obtain its latent representation at diffusion step $0$ via the encoder of variational autoencoder (VAE), denoted by $\mathbf{z}^{(src)}_0$. We then invert $\mathbf{z}^{(src)}_0$ to an intermediate diffusion step $t$ to obtain $\mathbf{z}^{(src)}_t$, which serves as a stable initialization for subsequent optimization. 


Let $\{\alpha_t\}_{t=0}^{T}$ denote the cumulative noise schedule used by DDIM, and let $\epsilon_{\theta}(\cdot)$ be the pretrained noise-prediction 3D U-Net. We adopt deterministic DDIM dynamics ($\eta=0$), so that both inversion and sampling are deterministic given the starting latent.

Starting from $\mathbf{z}_0^{(src)}$, DDIM inversion iteratively constructs noisier latents by
\begin{equation}
\begin{split}
\mathbf{z}^{(src)}_{t+1}
&=
\sqrt{\frac{\alpha_{t+1}}{\alpha_t}}\, \mathbf{z}^{(src)}_{t} \\
&+
\left(
\sqrt{\frac{1}{\alpha_{t+1}}-1}
-
\sqrt{\frac{1}{\alpha_t}-1}
\right)\epsilon_{\theta}\!\left(\mathbf{z}^{(src)}_{t},\, t\right),
\label{eq:ddim_inversion}
\end{split}
\end{equation}
yielding the intermediate latent $\mathbf{z}^{(src)}_t$ used to initialize $\mathbf{z}_{\mathrm{adv}}$ in Sec.~\ref{sec:method_opt}. Given any latent at step $t$ (in particular, the current $\mathbf{z}_{\mathrm{adv}}$ during optimization), deterministic DDIM sampling denoises it back to step $0$ via
\begin{equation}
\begin{split}
\mathbf{z}_{t-1}
&=
\sqrt{\frac{\alpha_{t-1}}{\alpha_t}}\, \mathbf{z}_{t} \\
&+
\left(
\sqrt{\frac{1}{\alpha_{t-1}}-1}
-
\sqrt{\frac{1}{\alpha_t}-1}
\right)\epsilon_{\theta}\!\left(\mathbf{z}_{t},\, t\right),
\label{eq:ddim_sampling}
\end{split}
\end{equation}
applied sequentially for $t,t-1,\ldots,1$ to obtain $\mathbf{z}_0$. The resulting $\mathbf{z}_0$ is then decoded by the VAE decoder to produce a continuous-valued sequence $\widetilde{\mathbf{x}}$, which is converted into a near-binary silhouette sequence by Sec.~\ref{sec:method_softbin}. Using an intermediate step $t$ provides a practical trade-off: inversion preserves coarse content from the source while leaving sufficient degrees of freedom for identity-level manipulation in the optimization loop.

\subsection{Differentiable Soft Binarization for Bimodal Silhouettes}
\label{sec:method_softbin}
Silhouette representations are expected to be bimodal, with clear foreground–background separation, whereas the denoised and VAE-decoded sequence $\widetilde{\mathbf{x}}$ is generally continuous-valued. During optimization, small gray-level deviations may appear near boundaries, weakening the silhouette prior and introducing artifacts. To encourage bimodality while preserving end-to-end differentiability, we apply a temperature-controlled sigmoid transformation.

Let $\widetilde{\mathbf{x}}\in[0,1]^{L\times H\times W}$ denote the continuous-valued sequence obtained after VAE decoding. We define the differentiable soft binarization as the element-wise mapping
\begin{equation}
\mathbf{x}^{(pro)}
=
\sigma\!\left(\frac{\widetilde{\mathbf{x}} - 0.5}{\tau}\right),
\label{eq:soft_binarization}
\end{equation}
where $\sigma(\cdot)$ is the sigmoid function, the threshold is fixed to $0.5$, and $\tau>0$ is a temperature parameter controlling sharpness. Smaller $\tau$ yields outputs closer to hard thresholding while producing steeper gradients near the decision boundary. We use $\tau=0.1$ as a robust default, which typically balances gradient stability and a strong bimodal tendency. The resulting $\mathbf{x}^{(pro)}$ is then used to compute identity-related objectives in Sec.~\ref{sec:method_opt}.

\subsection{Optimization Loop and Unified Objective}
\label{sec:method_opt}

We now describe how GaitProtector produces a protected gait by optimizing an intermediate diffusion latent initialized from inversion. Given $\mathbf{z}_t^{(src)}$ from Eq.~\eqref{eq:ddim_inversion}, we initialize $\mathbf{z}_{\mathrm{adv}} \leftarrow \mathbf{z}_t^{(src)}$ and iteratively refine $\mathbf{z}_{\mathrm{adv}}$ using identity-guided gradients. At each iteration, we (i) deterministically denoise $\mathbf{z}_{\mathrm{adv}}$ from step $t$ to $0$ using Eq.~\eqref{eq:ddim_sampling} to obtain $\mathbf{z}^{(pro)}_0$, (ii) apply VAE decoding to obtain $\widetilde{\mathbf{x}}^{(pro)}$, and (iii) apply Eq.~\eqref{eq:soft_binarization} to obtain $\mathbf{x}^{(pro)}$. For compactness, we denote this differentiable mapping by $\mathbf{x}^{(pro)}=\Phi\!\left(\mathbf{z}_{\mathrm{adv}}\right)$, where $\Phi(\cdot)$ composes deterministic DDIM denoising, VAE decoding, and differentiable soft binarization.

In practice, the exact recognition model used for evaluation may be unavailable. To reflect this black-box setting, we use an ensemble of $K$ surrogate gait feature extractors $\{\mathcal{G}_k\}_{k=1}^{K}$ to define identity similarity. For each $k$, we compute $\ell_2$-normalized embeddings
\begin{equation}
\mathbf{e}^{(i)}_k = \mathcal{G}_k\!\left(\mathbf{x}^{(i)}\right),
\quad i \in \{pro, src, tar\}.
\label{eq:embed_defs}
\end{equation}
and measure similarity via cosine similarity. Then our framework instantiates the dissimilarity $\mathcal{D}(\cdot,\cdot)$ in Sec.~\ref{sec:method_overview} using cosine distance (equivalently, negative cosine similarity) over this surrogate ensemble. Therefore, we define the impersonation (pull-to-target) objective by
\begin{equation}
\mathcal{L}_{\mathrm{imp}}
=
\frac{1}{K}\sum_{k=1}^{K}
\left(1-\mathrm{cos}\!\left(\mathbf{e}^{(pro)}_k,\mathbf{e}^{(tar)}_k\right)\right),
\label{eq:loss_imp_opt}
\end{equation}
and the obfuscation (repel-source) objective by
\begin{equation}
\mathcal{L}_{\mathrm{obf}}
=
\frac{1}{K}\sum_{k=1}^{K}
\mathrm{cos}\!\left(\mathbf{e}^{(pro)}_k,\mathbf{e}^{(src)}_k\right).
\label{eq:loss_obf_opt}
\end{equation}
Minimizing Eq.~\eqref{eq:loss_imp_opt} increases similarity to the target identity, while minimizing Eq.~\eqref{eq:loss_obf_opt} decreases similarity to the source identity. Importantly, the target term provides a semantic anchor. Obfuscation alone can be under-constrained and may drift toward identity-irrelevant distortions, whereas simultaneously attracting a specific target tends to yield more controlled and structurally plausible modifications.

Finally, we optimize $\mathbf{z}_{\mathrm{adv}}$ by minimizing the weighted sum
\begin{equation}
\min_{\mathbf{z}_{\mathrm{adv}}}\ \mathcal{L}
=
\lambda_{\mathrm{imp}}\,\mathcal{L}_{\mathrm{imp}}
+
\lambda_{\mathrm{obf}}\,\mathcal{L}_{\mathrm{obf}},
\label{eq:loss_total_opt}
\end{equation}
where $\lambda_{\mathrm{imp}}$ and $\lambda_{\mathrm{obf}}$ are hyperparameters to reflect the desired balance between impersonation and source suppression. Empirically, setting $\lambda_{\mathrm{imp}}$ in the range $[1,2]$ and $\lambda_{\mathrm{obf}}$ in the range $[0,0.3]$ provides a favorable privacy--quality trade-off.

\section{Experimental Settings}

\subsection{Datasets}
We evaluate privacy protection and visual quality on CASIA-B~\cite{CASIA-B}, and assess downstream clinical utility on Scoliosis1K~\cite{Scoliosis1K}.

\textbf{CASIA-B~\cite{CASIA-B}} contains 124 subjects captured under three walking conditions (\texttt{nm}, \texttt{bg}, \texttt{cl}) from 11 camera views. We follow the standard OpenGait~\cite{opengait} train--test split and conduct all experiments on the test subset.

For targeted protection, we designate two test identities (\texttt{102} and \texttt{103}) as targets and randomly select 15 other test identities as sources. Both source and target sequences are sampled from the OpenGait probe set (\texttt{nm-05}, \texttt{nm-06}, \texttt{bg-01}, \texttt{bg-02}, \texttt{cl-01}, \texttt{cl-02}). For each source identity, we construct 8 source--target pairs by randomly sampling a source sequence and pairing it with a target sequence, yielding 120 pairs in total.

For evaluation, protected sequences are treated as probes, while the gallery follows the standard OpenGait protocol and consists of test sequences under \texttt{nm-01} to \texttt{nm-04}.

\textbf{Scoliosis1K~\cite{Scoliosis1K}} is a large-scale clinical gait dataset for video-based scoliosis screening, containing over 1,000 subjects annotated with scoliosis-related clinical labels (negative, neutral, or positive). We conduct experiments on the test split defined in~\cite{Scoliosis1K}, which contains 784 silhouette sequences.

For targeted protection, we randomly select one neutral sequence as the target and sample 150 source sequences from the remaining test data across all clinical categories. Each source sequence is paired with the fixed target sequence. As identity labels are unavailable, privacy is evaluated at the sequence-level retrieval metrics, while downstream utility is assessed using a pretrained scoliosis classifier.

\vspace{-0.25em}
\subsection{Threat Model and Protocol}
We define the threat model and evaluation protocol used to assess gait de-identification under two main settings.

\textbf{Black-box evaluation:} We adopt a strict black-box threat model and evaluate privacy protection using three state-of-the-art silhouette-based gait recognition models: GaitBase~\cite{opengait}, DeepGaitV2~\cite{OpenGait_TPAMI}, and SwinGait~\cite{OpenGait_TPAMI}. To enforce backbone isolation, each experimental round uses two models as surrogate recognizers while reserving the third for evaluation (Table~\ref{tab:experimental_settings}). To further strengthen the black-box setting, surrogate models are pretrained on Gait3D~\cite{gait3d}, while the evaluation model is pretrained on CASIA-B, preventing information leakage across both backbone architectures and training datasets.

\begin{table}[t]
\centering
\caption{Experimental settings for black-box protection.}
\label{tab:experimental_settings}
\resizebox{\linewidth}{!}{%
\begin{tabular}{c|c|c}
\toprule
\multirow{2}{*}{\textbf{Setting ID}} & \textbf{Surrogate Models } & \textbf{Evaluation Model} \\
 & \textit{(Pre-trained on \textbf{Gait3D}~\cite{gait3d})} & \textit{(Pre-trained on \textbf{CASIA-B}~\cite{CASIA-B})} \\
\midrule
1 & DeepGaitV2~\cite{OpenGait_TPAMI} + SwinGait~\cite{OpenGait_TPAMI} & GaitBase~\cite{opengait} \\
2 & GaitBase~\cite{opengait} + SwinGait~\cite{OpenGait_TPAMI} & DeepGaitV2~\cite{OpenGait_TPAMI} \\
3 & GaitBase~\cite{opengait} + DeepGaitV2~\cite{OpenGait_TPAMI} & SwinGait~\cite{OpenGait_TPAMI} \\
\bottomrule
\end{tabular}%
}
\vspace{-1em}
\end{table}

\textbf{Adaptive adversary:} Although gait silhouettes are conceptually binary, real-world preprocessing often introduces small intermediate gray values. With $\epsilon=0.01$, approximately $2.8\%$ of pixels in CASIA-B and Scoliosis1K fall within $(\epsilon,1-\epsilon)$. Most existing gait recognizers do not enforce strict binarization, which allows methods relying on subtle pixel-level perturbations to affect recognition. To simulate a stronger adversary, we model an adaptive adversary that applies hard re-binarization to input silhouettes before evaluation, removing small gray-level perturbations and enforcing a strictly binary distribution. Under this threat model, effective privacy protection must rely on structurally meaningful modifications of silhouette shape and motion rather than imperceptible pixel noise.

\vspace{-0.25em}
\subsection{Baselines}
\label{sec:baselines}

All compared methods utilize embeddings and gradients from the same surrogate models and are assessed under a unified black-box protocol. We include one deformation-oriented non-diffusion baseline and three variants of our framework. Specifically, we implement a contour-based \emph{projected gradient descent (PGD)} baseline method that explicitly edits silhouette geometry (rather than pixel-level noise) to remain effective under hard re-binarization. Inspired by momentum-based projected adversarial optimization~\cite{dong2018boosting,madry2017towards}, it performs projected gradient updates on contour-localized degrees of freedom under an $\ell_\infty$ constraint and keeps the output strictly binary after each iteration. We further report \emph{Ours (VAE only)}, which optimizes only in the pretrained VAE latent space without 3D U-Net denoising; \emph{Ours (Obfuscation only)}, which removes the target-anchored term by setting $\mathcal{L}_{\mathrm{imp}}=0$; and \emph{Ours (Full)}, i.e., the complete \textbf{GaitProtector} framework.

\vspace{-0.25em}
\subsection{Evaluation Metrics}
\label{sec:metrics}
We evaluate gait de-identification from three complementary perspectives: privacy protection, visual quality, and downstream utility.

\textbf{Privacy:} On CASIA-B, we report \textit{Impersonation Success Rate (ISR)} and the change of Rank-1 accuracy before vs.\ after protection.
For each protected probe $\mathbf{x}^{(pro)}_i$, we compute its embedding $\mathcal{G}(\mathbf{x}^{(pro)}_i)$ using the evaluation model and retrieve the closest gallery sample under cosine distance. Let $\mathrm{Rank\text{-}1}(\mathbf{x})$ denote the identity of this Rank-1 match. ISR is:
\begin{equation}
\mathrm{ISR}
=
\frac{1}{N}
\sum_{i=1}^{N}
\mathbb{I}
\left[
\mathrm{Rank}\text{-}1\!\left(\mathbf{x}^{(pro)}_i\right)
=
y^{tar}_i
\right],
\end{equation}
where $N$ is the number of protected probes and $y^{tar}_i$ is the target identity. We also report how Rank-1 accuracy changes when replacing each source probe $\mathbf{x}^{(src)}_i$ with its protected version $\mathbf{x}^{(pro)}_i$ under the same evaluation protocol. 

On Scoliosis1K (without person-ID labels), we quantify privacy at the sequence level using retrieval-rank shifts. Specifically, we measure how a protected sequence is (i) drawn toward a designated target and (ii) repelled from its source in the embedding space of a gait recognizer, reporting the resulting rank changes before and after protection.

\textbf{Quality:} Static visual quality is evaluated frame-wise using \textit{Fr\'echet Image Distance (FID)}~\cite{heusel2017gans}, \textit{Peak Signal-to-Noise Ratio (PSNR)}, \textit{Structural Similarity (SSIM)}~\cite{wang2004image}, and \textit{Learned Perceptual Image Patch Similarity (LPIPS)}~\cite{zhang2018unreasonable}, while dynamic quality is assessed at the video level using \textit{Fr\'echet Video Distance (FVD)}~\cite{unterthiner2018towards}.

\textbf{Utility:} On Scoliosis1K, we evaluate downstream utility using a pretrained scoliosis classifier and report the performance drop on protected sequences.

\subsection{Implementation Details}
All sequences are converted to fixed-length inputs by deterministically selecting $30$ frames from the temporal center. Sequences shorter than $30$ frames are extended by cyclically repeating frames, while longer sequences are truncated to a contiguous center window of length $30$. DDIM inversion is performed with $T=20$ steps, initialized at diffusion step $t=3$. Latent optimization runs for $50$ iterations using AdamW with a learning rate of $0.1$. For \emph{GaitProtector (Full)}, we set $\lambda_{\mathrm{imp}}=1.5$ and $\lambda_{\mathrm{obf}}=0.1$. All other settings are identical across experiments, and all experiments are conducted on a single NVIDIA A100 GPU. On this setup, generating one protected 30-frame gait sequence takes roughly 3 minutes.

\section{Experimental Results}
\subsection{Results on CASIA-B}
\label{sec:results_casiab}

CASIA-B provides identity labels and a standard gallery/probe protocol, enabling direct evaluation of targeted impersonation (ISR) and source identity suppression (Rank-1 accuracy change), together with visual quality.

\subsubsection{Privacy Performance}
\label{sec:casiab_privacy}

Table~\ref{tab:privacy_casiab_per_backbone} reports privacy results under the adaptive adversary protocol, where inputs are hard re-binarized before evaluation.
\textit{GaitProtector (Full)} achieves the strongest targeted protection, reaching a mean ISR of $47.5\%$ across three black-box evaluators, while consistently reducing Rank-1 accuracy.
Compared with the contour-based PGD baseline, our method improves mean ISR by $+17.2$ points and maintains effective suppression under backbone and pretraining-data isolation.
Ablation studies confirm our design: removing diffusion denoising (\emph{VAE only}) substantially weakens both impersonation and suppression, whereas removing impersonation guidance (\emph{Obfuscation only}) drives Rank-1 accuracy down but yields zero impersonation success.

\begin{table*}[t]
\centering
\caption{Privacy results on CASIA-B: ISR and Rank-1 accuracy before/after protection for each evaluation backbone.}
\label{tab:privacy_casiab_per_backbone}
\resizebox{\linewidth}{!}{%
\begin{tabular}{l|cccc|cccccccc}
\toprule
\multirow{2}{*}{Method}
& \multicolumn{4}{c|}{ISR$\uparrow$ (\%)}
& \multicolumn{8}{c}{Rank-1 Accuracy (\%) $\downarrow$} \\
\cmidrule(lr){2-5}\cmidrule(lr){6-13}
& DeepGaitV2 & SwinGait & GaitBase & Mean
& \multicolumn{2}{c}{DeepGaitV2} & \multicolumn{2}{c}{SwinGait} & \multicolumn{2}{c}{GaitBase} & \multicolumn{2}{c}{Mean} \\
\cmidrule(lr){6-7}\cmidrule(lr){8-9}\cmidrule(lr){10-11}\cmidrule(lr){12-13}
&  &  &  &
& Before & After$\downarrow$ & Before & After$\downarrow$ & Before & After$\downarrow$ & Before & After$\downarrow$ \\
\midrule
Contour-based PGD
& \underline{31.7} & \underline{30.0} & \underline{29.2} & \underline{30.3}
& 89.6 & \underline{3.3} & 85.9 & \underline{1.7} & 86.1 & \textbf{10.0} & 87.2 & \textbf{5.0} \\

Ours (VAE only)
& 23.3 & 23.3 & 20.8 & 22.5
& 89.6 & 25.0 & 85.9 & 21.7 & 86.1 & 47.5 & 87.2 & 31.4 \\

Ours (Obfuscation only)
& 0 & 0 & 0 & 0
& 89.6 & \textbf{1.7} & 85.9 & \textbf{0.8} & 86.1 & \underline{20.8} & 87.2 & \underline{7.8} \\

Ours (Full)
& \textbf{56.7} & \textbf{53.3} & \textbf{32.5} & \textbf{47.5}
& 89.6 & 15.0 & 85.9 & 13.3 & 86.1 & 31.7 & 87.2 & 20.0 \\
\bottomrule
\end{tabular}%
}
\vspace{-0.9em}
\end{table*}

\subsubsection{Embedding Evidence}
\label{sec:casiab_tsne}
\begin{figure}[t]
  \centering
  \includegraphics[width=0.9\linewidth]{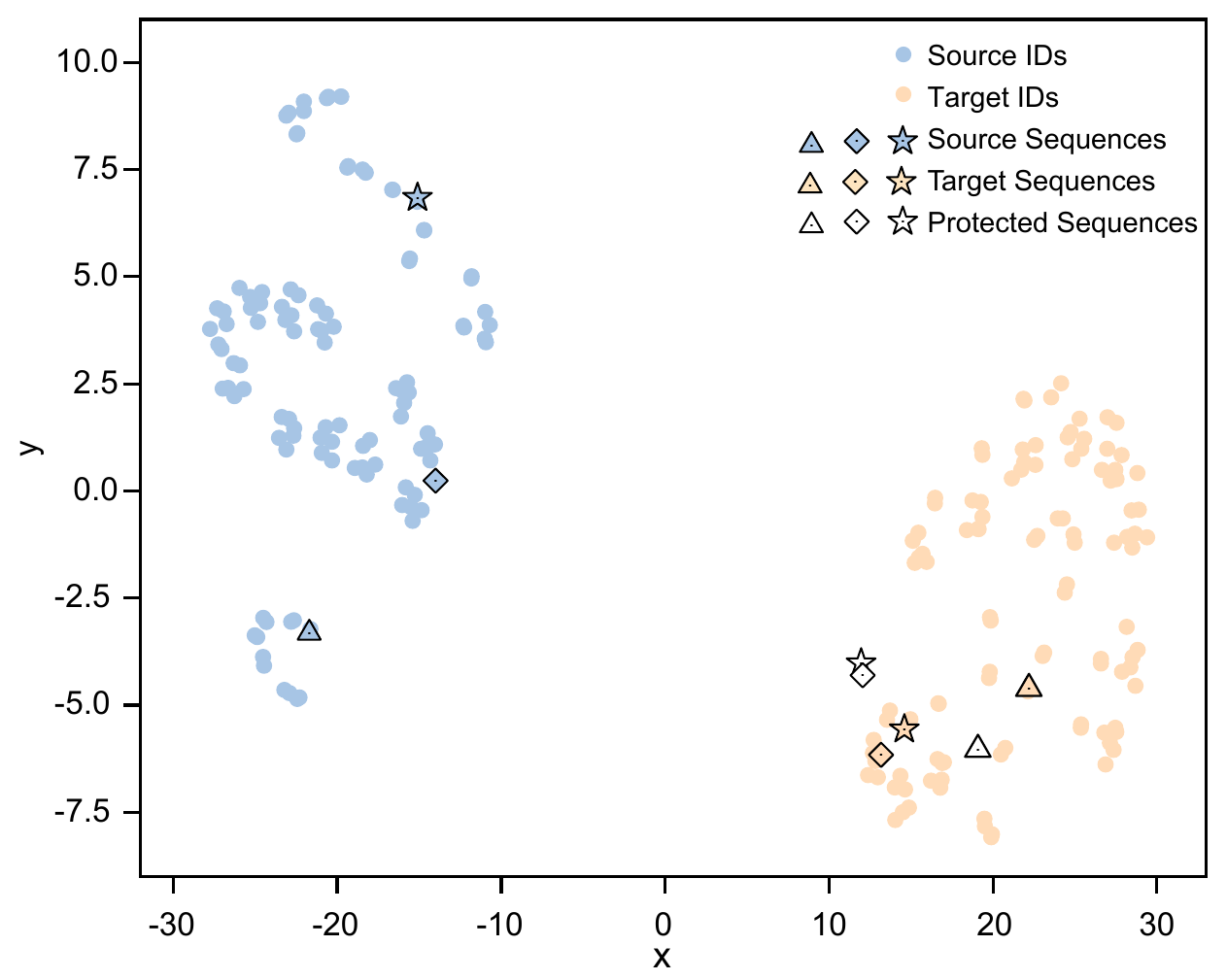}
  \vspace{-0.5em}
  \caption{\textbf{Embedding-space visualization on CASIA-B (t-SNE).}
  Blue/orange circles are embeddings from source/target IDs, forming identity clusters. Triangle/diamond/star indicate three example pairs; for each pair, the \textbf{white} marker is the protected sequence generated from its source.
  Protected embeddings consistently move away from the source side and toward the target side, reflecting obfuscation and impersonation.}
  \vspace{-1.3em}
  \label{fig:tsne}
\end{figure}

To complement rank-based metrics, we visualize how protection reshapes identity relations in feature space.
Fig.~\ref{fig:tsne} shows a 2D t-SNE~\cite{maaten2008visualizing} projection of embeddings extracted by \textbf{DeepGaitV2 pretrained on CASIA-B}.
Relative to the source clusters, protected samples move toward their designated targets, providing qualitative evidence for the intended \emph{push--pull} effect of the unified objective.

\subsubsection{Visual Quality}
\label{sec:casiab_quality}

We evaluate frame-wise fidelity by comparing protected frames with their aligned source frames, and assess temporal and video-level quality using FVD.
As reported in Table~\ref{tab:overall_mean_casiab}, our complete framework achieves the best overall quality, with the lowest FVD/FID and substantially improved LPIPS, indicating stronger temporal coherence and perceptual similarity.
In contrast, \emph{VAE only} yields severely degraded video-level quality (high FVD), highlighting the crucial role of the frozen diffusion prior in maintaining coherent spatiotemporal structure.

\begin{table}[t]
\centering
\caption{Visual quality on CASIA-B (mean over three evaluation backbones).}
\label{tab:overall_mean_casiab}
\resizebox{\linewidth}{!}{%
\begin{tabular}{l|cc|ccccc}
\toprule
Method
& \makecell{Mean\\ISR$\uparrow$}
& \makecell{Mean\\Rank-1$\downarrow$}
& FVD$\downarrow$
& FID$\downarrow$
& PSNR$\uparrow$
& SSIM$\uparrow$
& LPIPS$\downarrow$ \\
\midrule
Contour-based PGD
& \underline{30.3}
& \textbf{5.0}
& \underline{144.57}
& \underline{42.68}
& 14.86
& \textbf{0.810}
& 0.137 \\
Ours (VAE only)
& 22.5
& 31.4
& 417.97
& 82.69
& \underline{19.00}
& 0.733
& \underline{0.131} \\
Ours (Obfuscation only)
& 0
& \underline{7.8}
& 353.89
& 63.77
& 17.13
& 0.735
& 0.134 \\
Ours (Full)
& \textbf{47.5}
& 20.0
& \textbf{128.66}
& \textbf{34.93}
& \textbf{19.46}
& \underline{0.805}
& \textbf{0.088} \\
\bottomrule
\end{tabular}%
}
\vspace{-0.5em}
\end{table}

\subsubsection{Qualitative Comparison}
\label{sec:casiab_qual}

\begin{figure*}[t]
  \centering
  \includegraphics[width=0.9\linewidth]{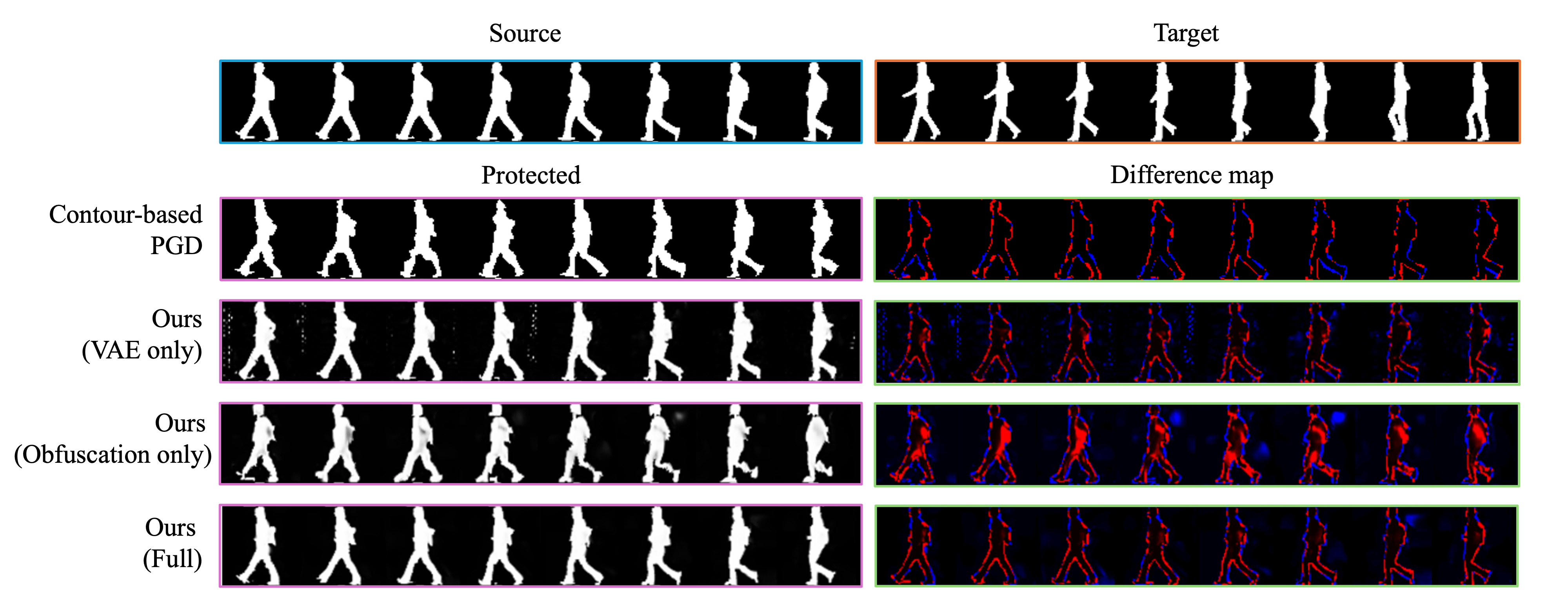}
  \vspace{-0.5em}
  \caption{\textbf{Qualitative comparison on CASIA-B.}
  We show a representative source–target pair (top) alongside protected sequences generated by different methods (left) and per-frame difference maps relative to the source (right). In these maps, pixels that become \emph{whiter} are shown in \textcolor{blue}{blue}, and pixels that become \emph{blacker} in \textcolor{red}{red}, with changes amplified $10\times$ for clarity.}
  \label{fig:baselines}
  \vspace{-0.5em}
\end{figure*}

Fig.~\ref{fig:baselines} provides a qualitative comparison on a representative source--target pair.
The contour-based PGD baseline yields sharp and boundary-localized edits, resulting in less human-like shapes.
\emph{Ours (VAE only)} frequently exploits background pixels (visible as scattered changes outside the body region in the difference maps), suggesting that optimizing only the VAE latent can introduce off-manifold artifacts without a strong spatiotemporal prior.
\emph{Ours (Obfuscation only)} is further under-constrained: without a target anchor, the optimization can drift and ``pollute'' the silhouette stream with broader, less structured modifications, which are not clearly aligned with plausible gait geometry.
In contrast, \textbf{GaitProtector (Full)} produces more coherent and visually plausible adjustments: while soft binarization may still induce minor background changes, the dominant edits concentrate on subtle shape refinements around the body contour, largely preserving the source's main trunk structure and motion pattern.

\subsubsection{Robustness to Re-binarization}
\label{sec:casiab_robust}

Table~\ref{tab:robustness_one_row} evaluates robustness against the adaptive adversary by comparing performance \emph{without} vs.\ \emph{with} adversarial re-binarization at test time.
Despite removing intermediate gray values, re-binarization causes only a small drop in targeted impersonation (ISR: $52.5\%\!\rightarrow\!47.5\%$) and a mild increase in post-protection identifiability (Rank-1 after: $17.8\%\!\rightarrow\!20.0\%$).
This indicates that GaitProtector's privacy gains do not primarily rely on fragile gray-level perturbations and largely persist under strict binary enforcement.

\begin{table}[t]
\centering
\caption{Robustness of \textbf{GaitProtector} on CASIA-B: mean ISR and Rank-1 (after) \emph{without} vs.\ \emph{with} adversarial re-binarization.}
\label{tab:robustness_one_row}
\resizebox{\linewidth}{!}{%
\begin{tabular}{lcccccc}
\toprule
Method
& \makecell{ISR$\uparrow$\\w/o re-bin.}
& \makecell{ISR$\uparrow$\\w/ re-bin.}
& \makecell{$\Delta$ISR}
& \makecell{Rank-1$\downarrow$\\w/o re-bin.}
& \makecell{Rank-1$\downarrow$\\w/ re-bin.}
& \makecell{$\Delta$Rank-1}
\\
\midrule
Ours (Full)
& 52.5
& 47.5
& -5.0
& 17.8
& 20.0
& +2.2
\\
\bottomrule
\end{tabular}%
}
\vspace{-1.3em}
\end{table}

\subsection{Results on Scoliosis1K}

Since Scoliosis1K lacks person-ID labels, we evaluate privacy at the \emph{sequence level} using retrieval-rank shifts in the embedding space of a gait recognizer. 
Specifically, we use DeepGaitV2 (pretrained on Gait3D~\cite{gait3d}) to embed each sequence, treat each \emph{source} (or its \emph{protected} output) as a query, and retrieve against the remaining test sequences as the gallery.
We quantify (i) \emph{pull-to-target} by the rank of the designated target sequence, and (ii) \emph{push-from-source} by the rank of the original source sequence.
We additionally report video and frame quality (FVD and FID) and downstream clinical utility using a pretrained ScoNet~\cite{Scoliosis1K} classifier.

\subsubsection{Sequence-level Proxy Privacy}
\label{sec:scoliosis_privacy}

Table~\ref{tab:scoliosis_seqlevel} summarizes sequence-level privacy results and reports two complementary effects: (i) \emph{pull-to-target}, measured by the target's rank when a protected sequence is used as the probe (lower is better), and (ii) \emph{push-from-source}, measured by the source's rank when querying with its protected output (higher is better).
\textbf{GaitProtector (Full)} achieves substantially stronger pull-to-target than the contour-based PGD baseline (median target rank: $92$ vs.\ $243$) while maintaining comparable source suppression, demonstrating effective identity redirection without compromising overall privacy.

\begin{table}[t]
\centering
\caption{Sequence-level proxy privacy, quality, and utility on Scoliosis1K.}
\label{tab:scoliosis_seqlevel}
\resizebox{\linewidth}{!}{%
\begin{tabular}{l|cc|cc|cc|cc|cc}
\toprule
\multirow{2}{*}{Method}
& \multicolumn{2}{c|}{\makecell{Source probe\\Target rank$\downarrow$}}
& \multicolumn{2}{c|}{\makecell{Protected probe\\Target rank$\downarrow$}}
& \multicolumn{2}{c|}{\makecell{Protected probe\\Source rank$\uparrow$}}
& \multicolumn{2}{c|}{Quality$\downarrow$}
& \multicolumn{2}{c}{Utility$\uparrow$} \\
\cmidrule(lr){2-3}\cmidrule(lr){4-5}\cmidrule(lr){6-7}\cmidrule(lr){8-9}\cmidrule(lr){10-11}
& Mean & Median
& Mean & Median
& Mean & Median
& FVD & FID
& \makecell{Acc.\\(source)} & \makecell{Acc.\\(protected)} \\
\midrule
Contour-based PGD
& 325.31 & 338.00
& 255.11 & 243.00
& 284.07 & \textbf{251.00}
& 129.62 & 39.04
& 91.4 & 59.3 \\
Ours (Full)
& 325.31 & 338.00
& \textbf{99.57} & \textbf{92.00}
& \textbf{287.72} & 220.50
& \textbf{98.27} & \textbf{35.13}
& 91.4 & \textbf{74.2} \\
\bottomrule
\end{tabular}%
}

\vspace{-1.1em}
\end{table}

\subsubsection{Visualization of Rank Shifts}
\label{sec:scoliosis_rankvis}

\begin{figure}[t]
  \centering
  \includegraphics[width=0.9\linewidth]{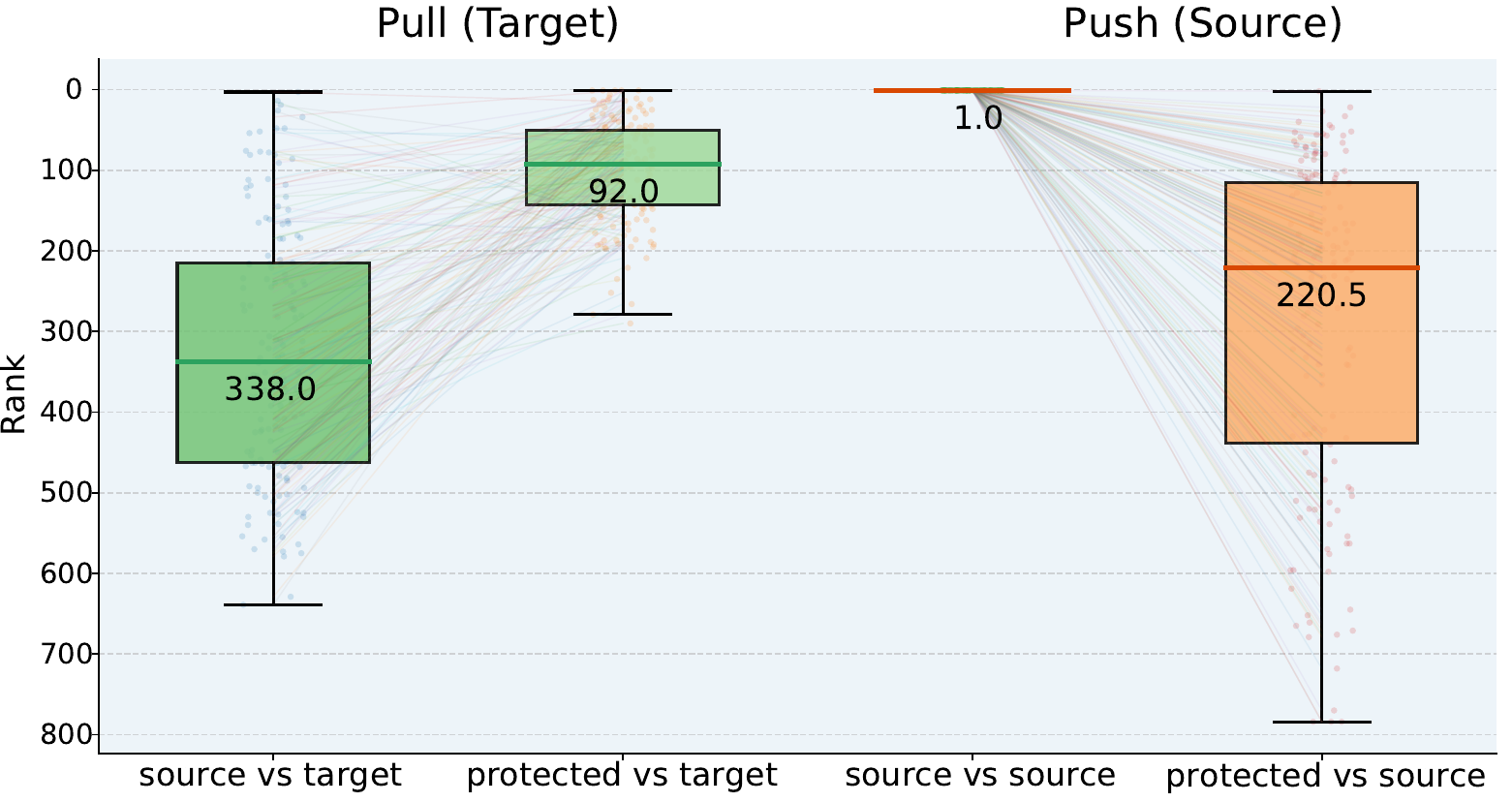}
  \vspace{-0.3em}
  \caption{\textbf{Sequence-level rank shifts on Scoliosis1K.}
  Each box summarizes ranks over 150 source sequences (rank $1$ is best; smaller ranks appear higher on the axis).
  The left two columns show the rank position of the \emph{designated target} when using the \emph{source} sequence (``source vs target'') as the probe versus the \emph{protected} output (``protected vs target'') as the probe.
  The right two columns show the rank of the \emph{original source} when probing with the \emph{source} itself (``source vs source'', which is always $1$) versus the \emph{protected} output (``protected vs source'').
  Points indicate individual samples, and faint lines connect each source to its protected counterpart.}

  \label{fig:scoliosis_rank}
  \vspace{-0.5em}
\end{figure}

Fig.~\ref{fig:scoliosis_rank} visualizes rank changes for all source sequences.
On the left (\emph{Pull}), protected outputs elevate the designated target in the retrieval list, while on the right (\emph{Push}), they substantially degrade retrieval of the original source.
The consistent shifts across paired lines indicate that the privacy effect is not driven by a few outliers but holds broadly effective across the test sources.

\subsubsection{Quality and Utility Preservation}
\label{sec:scoliosis_utility}

Beyond privacy, Table~\ref{tab:scoliosis_seqlevel} shows that \textbf{GaitProtector (Full)} produces higher-fidelity gait videos and better preserves downstream clinical utility than contour-based PGD.
In particular, scoliosis classifier accuracy drops from $91.4$ on source sequences to $74.2$ on protected sequences, whereas contour-based PGD causes a larger degradation to $59.3$.
This gap is consistent with the fact that scoliosis screening is structure-sensitive: retaining the dominant body/trunk geometry and temporally consistent motion patterns is crucial for preserving diagnostic cues.

\begin{figure}[t]
  \centering
  \includegraphics[width=0.9\linewidth]{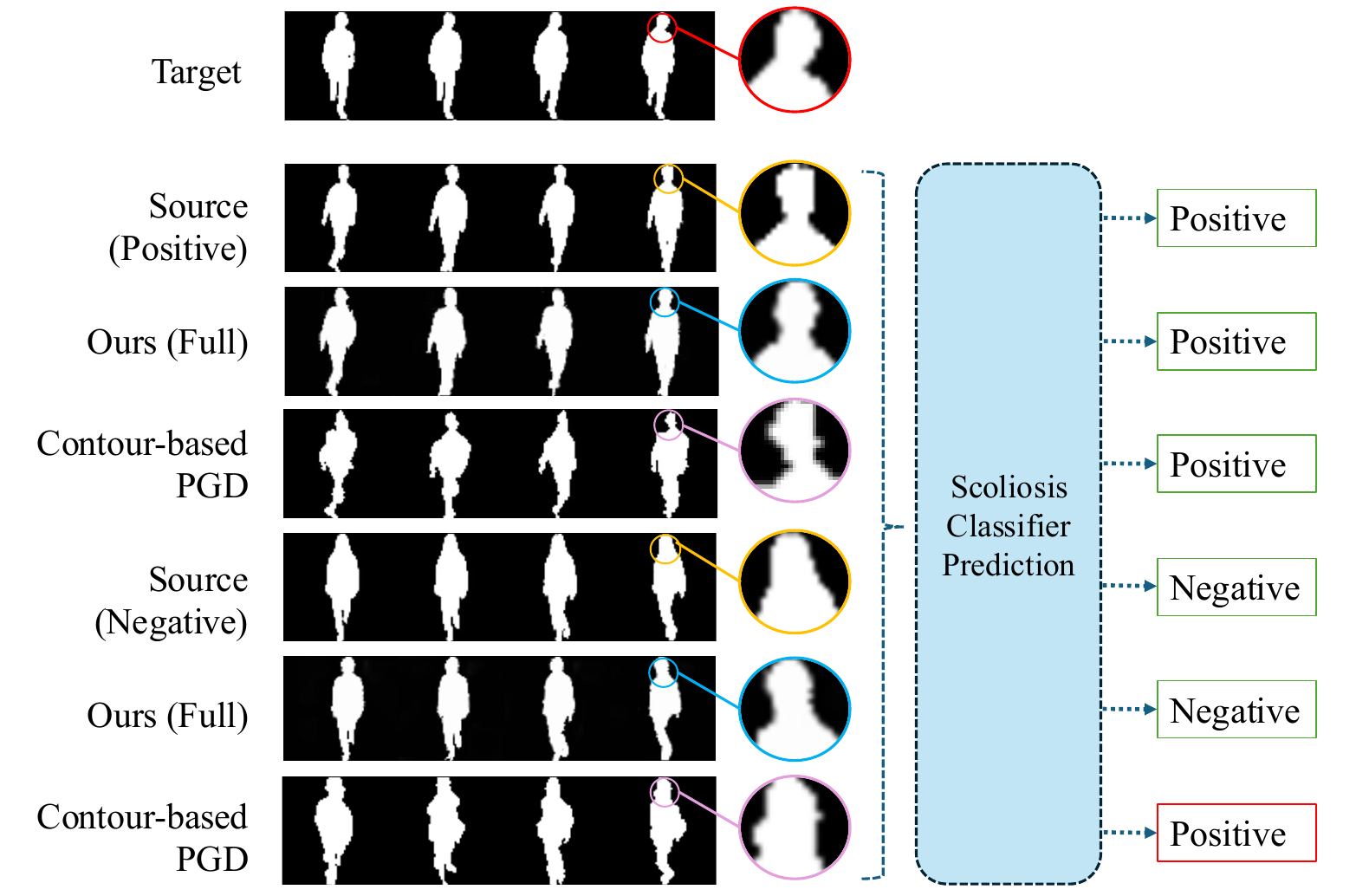}
  \vspace{-0.3em}
  \caption{\textbf{Qualitative utility preservation on Scoliosis1K.}
We show one designated target and two source gait sequences (one positive and one negative), together with protected outputs from our framework and contour-based PGD.
Insets magnify the head region to highlight identity-related shape changes. The right panel shows predictions from a pretrained scoliosis classifier.
Contour-based PGD yields less natural silhouettes; in the negative example, it flips the classifier decision, whereas GaitProtector better preserves global structure, motion, and the original prediction.}
  \label{fig:ScoliosisUtility}
  \vspace{-0.9em}
\end{figure}

Fig.~\ref{fig:ScoliosisUtility} provides a qualitative comparison.
Across both the positive and negative source gait sequences, contour-based PGD tends to introduce jagged, non-anatomical silhouette artifacts and less coherent shape evolution over time, even though the positive case remains correctly classified.
In contrast, GaitProtector largely preserves the source gait's global spatial structure (especially the trunk) and its temporal motion pattern, while making more subtle, localized shape adjustments that are sufficient for identity manipulation.
Notably, the head region (shown in the magnified insets) becomes subtly more similar to the designated target after protection, illustrating how our target-anchored objective steers identity-sensitive details toward the target without disrupting structure-critical cues for clinical prediction.

\section{Discussion}

\subsection{Why Impersonation Regularizes De-ID}
Our framework formulates gait de-identification as the joint effect of obfuscation and impersonation, rather than identity suppression alone. Purely repulsive objectives are inherently under-constrained and may drift toward identity-irrelevant distortions. Introducing an explicit impersonation target provides a semantic anchor in the surrogate embedding spaces, which, together with the frozen diffusion prior, biases optimization toward more plausible gait transformations. We therefore interpret impersonation as an empirical regularizer rather than as a formal guarantee of staying on the gait manifold. While this idea is related to targeted anonymization in face privacy, gait de-identification poses distinct challenges. Gait identity is encoded primarily through shape-level structure and spatiotemporal dynamics rather than appearance. In our setting, impersonation regularizes both static body geometry and motion patterns over time, which is important for preserving coherent gait dynamics while enabling controllable de-identification.

\subsection{Target Selection and Mismatch Sensitivity}
The effectiveness of impersonation depends on the relationship between source and target sequences. Mismatches in walking condition, camera view, or both lead to lower impersonation success rates and degraded visual quality, suggesting that current surrogate gait recognition models do not fully disentangle identity from nuisance factors such as viewpoint and condition. This implies that more identity-invariant gait representations could further benefit de-identification. In practice, selecting targets with similar conditions or views, or enforcing matching constraints, improves stability and quality. Broader temporal-sequence literature likewise suggests that robustness to incomplete observations and multi-view mismatch may be useful for future gait identity models~\cite{11002729,yu2025merlin}, although adapting such ideas to gait de-identification remains open. More generally, this sensitivity highlights a limitation of current representations, as fully unconstrained any-to-any gait de-identification remains challenging.

\subsection{Computation and Deployment Efficiency}
Although GaitProtector is training-free, its current optimization loop remains computationally demanding. This cost is acceptable for offline privacy-preserving dataset release, but it remains a limitation for interactive deployment and large-scale batch processing. This is especially relevant if privacy-preserving motion editing is to become an interactive component in broader controllable generation systems~\cite{li2026comprehensive}. A practical path forward is to reduce the cost of denoising and inversion through fewer DDIM steps, lighter latent-video backbones, and deployment-oriented compression or distillation of video diffusion models~\cite{he2022lvdm,blattmann2023align,feng2025q,feng2025s,li2026distilling}. More broadly, robustness-aware distillation and explicit multiobjective optimization may also help balance privacy, utility, and efficiency in future systems~\cite{li2025mmt,dongrobust,dong2026allies,11214262}. In addition, recent controllable video editing formulations that reduce dependence on costly inversion suggest another direction for lowering end-to-end latency~\cite{zhang2025visual}.

\subsection{Ethical Considerations and Intended Use}
GaitProtector is intended as a defensive tool for privacy protection rather than for real-world impersonation. The impersonation objective is used only to regularize the optimization and stabilize the protected output. In practical deployment, target anchors should preferably be consented prototypes or synthesized templates rather than unsuspecting real individuals, and protected outputs should be audited for both residual source identifiability and unintended target misattribution before release. More broadly, privacy leakage, controllability, and misuse concerns increasingly span multiple AI settings, including language-model inference, interaction understanding, and embodied manipulation~\cite{chu2025selective,yang2026unihoi,YANG2026114166}. The intended use of this work is controlled privacy-preserving data sharing and analysis, especially for benchmark release, secondary analysis, and cross-institution model development in sensitive gait applications.

\section{CONCLUSIONS}

We have introduced GaitProtector, a training-free gait de-identification framework that exploits pretrained 3D diffusion priors for controlled identity manipulation in silhouette-based gait sequences. By jointly enforcing obfuscation and impersonation in latent space, GaitProtector achieves effective privacy protection while preserving structurally meaningful gait dynamics. Experimental results demonstrate strong identity suppression, high visual fidelity, and maintained downstream clinical utility, as well as robustness under adaptive threat models.

{\small
\bibliographystyle{ieee}
\bibliography{egbib}
}

\end{document}